# Belief Update in CLG Bayesian Networks With Lazy Propagation

**Anders L Madsen**
HUGIN Expert A/S
Gasværksvej 5
9000 Aalborg, Denmark
Anders.L.Madsen@hugin.com

## Abstract

In recent years Bayesian networks (BNs) with a mixture of continuous and discrete variables have received an increasing level of attention. We present an architecture for exact belief update in Conditional Linear Gaussian BNs (CLG BNs). The architecture is an extension of lazy propagation using operations of Lauritzen & Jensen [6] and Cowell [2]. By decomposing clique and separator potentials into sets of factors, the proposed architecture takes advantage of independence and irrelevance properties induced by the structure of the graph and the evidence. The resulting benefits are illustrated by examples. Results of a preliminary empirical performance evaluation indicate a significant potential of the proposed architecture.

## 1 INTRODUCTION

The framework of BNs is an efficient knowledge representation for reasoning under uncertainty [12, 3, 4]. Traditionally, the variables of a BN are assumed to be either discrete or continuous. In recent years BNs with a mixture of continuous and discrete variables have received an increasing level of attention. Here exact belief update in CLG BNs is considered.

Extending the class of BNs containing discrete (or continuous) variables only to the class of BNs containing both discrete and continuous variables is not simple. The work by Pearl [12] on BNs containing continuous variables imposed three constraints on the variables in the network. The interaction between variables is linear, the sources of uncertainty are Gaussian distributed and uncorrelated, and the causal network is singly connected. Later, Shachter & Kenley [14] described how to solve Gaussian influence diagrams under similar constraints, but allowing multiple connected causal networks.

Lauritzen [5] presents a scheme for modeling and exact belief update in CLG BNs. This scheme is more general than the scheme proposed by Pearl. The conditional distribution of the continuous variables given the discrete variables is assumed to be multivariate Gaussian. Only continuous variables which are linear additively Gaussian distributed are considered. The asymmetry between continuous and discrete variables induces a number of constraints on the model specification and the inference structure. Using a similar approach Chang & Fung [1] extend the SPI algorithm [13] to solve arbitrary queries against CLG BNs.

The Lauritzen [5] architecture is known to suffer from problems causing numerical instability. For this reason the architecture was later revised by Lauritzen & Jensen [6] in order to improve the numerical stability of belief update. Recently, Cowell [2] introduced an alternative architecture for belief update based on message passing in an elimination tree using the arc-reversal operation of Shachter & Kenley [14] (referred to as the EXCHANGE operation). By performing message passing in an elimination tree the need for complex matrix operations is eliminated.

We introduce a new architecture for belief update in CLG BNs. The architecture is an extension of lazy propagation [10] based on operations introduced by Lauritzen & Jensen [6] and Cowell [2]. Belief update proceeds by message passing in a strong junction tree structure where messages are computed using arc-reversal and EXCHANGE operations. The EXCHANGE operation is extended to eliminate discrete variables by arc-reversal. Variables are eliminated using a sequence of EXCHANGE operations and barren variable removals. Posterior marginal distributions are computed using EXCHANGE and PUSH [6] operations.

We investigate the computational efficiency of the proposed architecture by comparing its performance on a number of randomly generated CLG BNs with the performance of the Lauritzen & Jensen [6] architecture as implemented in the HUGIN Decision Engine, i.e., the inference engine of the HUGIN tools. Furthermore, we analyze the performance of various steps of belief update such as computing posterior distributions.

## 2 PRELIMINARIES AND NOTATION

### 2.1 CLG BAYESIAN NETWORK

A CLG BN $\mathcal{N} = (\mathcal{X}, \mathcal{G}, \mathcal{P}, \mathcal{F})$ over variables $\mathcal{X}$ consists of an acyclic, directed graph (DAG) $\mathcal{G} = (V, E)$, a set of conditional probability distributions $\mathcal{P} = \{P(X | \pi(X)) : X \in \Delta\}$, and a set of CLG density functions $\mathcal{F} = \{f(Y | \pi(Y)) : Y \in \Gamma\}$ where $\Delta$ is the set of discrete variables and $\Gamma$ is the set of continuous variables such that $\mathcal{X} = \Delta \cup \Gamma$. The vertices $V$ of $\mathcal{G}$ correspond one to one with the variables of $\mathcal{X}$. CLG BN $\mathcal{N}$ induces a multivariate normal mixture density over $\mathcal{X}$ on the form:

$$P(\Delta) \cdot f(\Gamma | \Delta) = \prod_{X \in \Delta} P(X | \pi(X)) \cdot \prod_{Y \in \Gamma} f(Y | \pi(Y)),$$

where $\pi(X)$ is the set of variables corresponding to the parents of the vertex representing $X$ in $G$.

Let $Y \in \Gamma$ with $I = \pi(Y) \cap \Delta$ and $Z = \pi(Y) \cap \Gamma$, then $Y$ has a CLG distribution if:

$$\mathcal{L}(Y | I = i, Z = z) = N(\alpha(i) + \beta(i) z, \sigma^2(i)), \quad (1)$$

where the mean value of $Y$ depends linearly on the values of the continuous parent variables $Z$, while the variance is independent of $Z$. In (1), $\alpha(i)$ is a table of real numbers, $\beta(i)$ is a table of $|Z|$-dimensional vectors, and $\sigma^2(i)$ is a table of non-negative values.

Evidence on a variable $X \in \Delta$ is assumed to be an instantiation, i.e., $X = x$. Evidence on a variable $Y \in \Gamma$ is an assignment of a value $y$ to $Y$, i.e., $Y = y$. We let $\epsilon_\Delta$ and $\epsilon_\Gamma$ denote the set of evidence on variables of $\Delta$ and $\Gamma$, respectively, such that $\epsilon = \epsilon_\Delta \cup \epsilon_\Gamma$.

**Definition 2.1 [Barren Variable]**
A variable $X$ is a *barren variable* w.r.t. a set of variables $T$, evidence $\epsilon$, and DAG $\mathcal{G}$, if $X \notin T$, $X \notin \epsilon$ and $X$ only has barren descendants in $\mathcal{G}$ (if any).

### 2.2 THE EXCHANGE OPERATION

Let $Y, Z \in \Gamma$ with parent sets $\pi(Z) = \{Z_1, \ldots, Z_n\} \subseteq \Gamma$ and $\pi(Y) = \{Z, Z_1, \ldots, Z_n\} \subseteq \Gamma$ such that:

$$Y | Z, Z_1, \ldots, Z_n \sim N(\alpha_Y + \beta_Z Z + \sum_{i=1}^{n} \beta_i Z_i, \sigma_Y^2),$$

$$Z | Z_1, \ldots, Z_n \sim N(\alpha_Z + \sum_{i=1}^{n} \delta_i Z_i, \sigma_Z^2).$$

The EXCHANGE operation is essentially Bayes' theorem. It converts the above pair of distributions such that $Y$ becomes a parent of $Z$ in the domain graph spanned by the two distributions maintaining the same joint probability density function of the original pair [2]. Graphically speaking the EXCHANGE operation corresponds to arc-reversal in the domain graph. The distribution of $Y$ after EXCHANGE is:

$$Y | Z_1, \ldots, Z_n \sim$$
$$N(\alpha_Y + \beta_Z \alpha_Z + \sum_{i=1}^{n} (\beta_i + \beta_Z \delta_i) Z_i, \sigma_Y^2 + \beta_Z^2 \sigma_Z^2),$$

while the distribution of $Z$ is (Cowell [2] considers three different cases depending on the values of $\sigma_Y^2$ and $\sigma_Z^2$ that are mathematical limits of this case):

$$Z | Y, Z_1, \ldots, Z_n \sim N \left( \frac{\rho}{\sigma_Y^2 + \beta_Z^2 \sigma_Z^2}, \frac{\sigma_Z^2 \sigma_Y^2}{\sigma_Y^2 + \beta_Z^2 \sigma_Z^2} \right),$$

where

$$\rho = \alpha_Z \sigma_Y^2 - \alpha_Y \beta_Z \sigma_Z^2 + \sum_{i=1}^{n} (\delta_i \sigma_Y^2 - \beta_i \beta_Z \sigma_Z^2) Z_i + \beta_Z \sigma_Z^2 Y.$$

It is straightforward to extend the EXCHANGE operation to handle discrete variables. Let $X_i, X_j \in \Delta$ with parent sets $\pi(X_i) = \{X_1, \ldots, X_n\} \subseteq \Delta$ and $\pi(X_j) = \{X_i, X_1, \ldots, X_n\} \subseteq \Delta$ such that $p(X_j | X_i, X_1, \ldots, X_n)$ and $p(X_i | X_1, \ldots, X_n)$ are the corresponding probability potentials of $X_i$ and $X_j$, respectively. In the discrete case the EXCHANGE operation is also essentially Bayes' theorem. That is, the EXCHANGE operation converts the above pair of potentials such that $X_j$ becomes a parent of $X_i$ in the domain graph spanned by the two potentials maintaining the same joint probability potential of the original pair:

$$p(X_j | X_1, \ldots, X_n) = $$
$$\sum_{X_i} p(X_j | X_i, X_1, \ldots, X_n) p(X_i | X_1, \ldots, X_n),$$
$$p(X_i | X_j, X_1, \ldots, X_n) = $$
$$\frac{p(X_j | X_i, X_1, \ldots, X_n) p(X_i | X_1, \ldots, X_n)}{p(X_j | X_1, \ldots, X_n)}.$$

Graphically speaking the EXCHANGE operation corresponds to arc-reversal in the domain graph.

By construction it is never necessary to apply the EXCHANGE operation to a pair of mixed variables (i.e., a continuous and a discrete variable). Also, prior to applying the EXCHANGE operation on a pair of adjacent variables we make sure that the two variables share the same set of parents. This is achieved by straightforward domain extensions.

## 2.3 THE STRONG JUNCTION TREE

Belief update is performed by message passing in a strong junction tree $\mathcal{T} = (\mathcal{C}, \mathcal{S})$ with cliques $\mathcal{C}$, separators $\mathcal{S}$ and strong root $R \in \mathcal{C}$. $\mathcal{T}$ has the property that for all adjacent cliques $A$ and $B$ with $A$ closer to $R$ than $B$, it holds that $S = A \cap B \subseteq \Delta$ or $B \setminus A \subseteq \Gamma$. Let $A$ and $B$ be adjacent cliques with $A$ closer to $R$ than $B$ and such that $S = A \cap B$. Then, $A$ is referred to as the *parent clique* of $B$ (denoted $\pi_\mathcal{C}(B)$) and $S$ is referred to as the *parent separator* of $B$ (denoted $\pi_\mathcal{S}(B)$).

A clique $C \in \mathcal{C}$ is referred to as a *boundary clique* if $C \cap \Gamma \neq \emptyset$ and either $B \subseteq \Delta$ or $B \cap \Gamma$ is instantiated by evidence where $B = \pi_\mathcal{C}(C)$. Let $\mathrm{bd}(\mathcal{C})$ denote the set of boundary cliques.

# 3 LAZY PROPAGATION

A junction tree for a discrete BN is by construction *wide enough* to support the computation of any posterior marginal given any subset of evidence. The junction tree is, however, often too wide to take advantage of independence properties induced by evidence. Lazy propagation aims at taking advantage of independence and irrelevance properties induced by evidence in a Shenoy-Shafer message passing scheme [10]. In Lazy propagation belief update proceeds by message passing in a junction tree maintaining decompositions of clique and separator potentials.

## 3.1 POTENTIALS AND OPERATIONS

**Definition 3.1 [Potential]**
A *potential* on $W \subseteq \mathcal{X}$ is a pair $\pi_W = (\mathcal{P}, \mathcal{F})$ where $\mathcal{P}$ is a set of (discrete) probability potentials on subsets of $W$ and $\mathcal{F}$ is a set of probability density functions on subsets of $W \cap \Gamma$ conditional on subsets of $W \cap \Delta$.

Elements of $\mathcal{P}$ are referred to as *factors* and elements of $\mathcal{F}$ as *density functions* (or *densities*). Furthermore, we call a potential $\pi_W$ vacuous if $\pi_W = (\emptyset, \emptyset)$. The vacuous potential is denoted $\pi_\emptyset$.

**Definition 3.2 [Combination]**
The *combination* of two potentials $\pi_{W_1} = (\mathcal{P}_1, \mathcal{F}_1)$ and $\pi_{W_2} = (\mathcal{P}_2, \mathcal{F}_2)$ denotes the potential on $W_1 \cup W_2$ given by $\pi_{W_1} \otimes \pi_{W_2} = (\mathcal{P}_1 \cup \mathcal{P}_2, \mathcal{F}_1 \cup \mathcal{F}_2)$.

Notice that potential combination is set union.

**Definition 3.3 [Contraction]**
The contraction $c(\pi_W)$ of a potential $\pi_W = (\mathcal{P}, \mathcal{F})$ is the non-negative function on $W$ given by:

$$c(\pi_W) = \prod_{p \in \mathcal{P}} p \cdot \prod_{f \in \mathcal{F}} f.$$

We define the contraction of $\pi_\emptyset$ as $c(\pi_\emptyset) = 1$.

**Definition 3.4 [Projection]**
The *projection* of a potential $\pi_W = (\mathcal{P}_W, \mathcal{F}_W)$ to a subset $U \subseteq W$ denotes the potential $\pi_U = \pi_W^{\downarrow U} = (\mathcal{P}_U, \mathcal{F}_U)$ on $U$ obtained by performing a sequence of EXCHANGE operations and barren variable removals eliminating variables of $W \setminus U$.

In projection continuous variables are eliminated before discrete variables. Notice that the head of any factor or density will consists of a single variable or a single piece of evidence. If a variable $X$ is barren, then $X$ and its factor or density may be removed without further computation.

## 3.2 INITIALIZATION

The first step in initialization of $\mathcal{T} = (\mathcal{C}, \mathcal{S})$ is to associate $\pi_\emptyset$ with each clique $C \in \mathcal{C}$. Next, for each $X \in \Delta$, we assign $P(X | \pi(X)) \in \mathcal{P}$, to the clique $C$ closest to $R$ such that $\mathrm{fa}(X) \subseteq C$ where $\mathrm{fa}(X) = \pi(X) \cup \{X\}$. Similarly, for each $Y \in \Gamma$. After initialization each clique $C$ holds a potential $\pi_C = (\mathcal{P}, \mathcal{F})$. The joint potential $\pi_\mathcal{X}$ on $\mathcal{T} = (\mathcal{C}, \mathcal{S})$ is therefore:

$$\pi_\mathcal{X} = \bigotimes_{C \in \mathcal{C}} \pi_C = \left( \bigcup_{X \in \Delta} \{P(X | \pi(X))\}, \bigcup_{Y \in \Gamma} \{f(Y | \pi(Y))\} \right).$$

The contraction of the joint potential $\pi_\mathcal{X}$ is:

$$c(\pi_\mathcal{X}) = c(\bigotimes_{C \in \mathcal{C}} \pi_C) = \prod_{X \in \Delta} P(X | \pi(X)) \cdot \prod_{Y \in \Gamma} f(Y | \pi(Y)).$$

Evidence $\epsilon_\Delta$ is inserted as part of initialization while evidence $\epsilon_\Gamma$ is inserted during message passing.

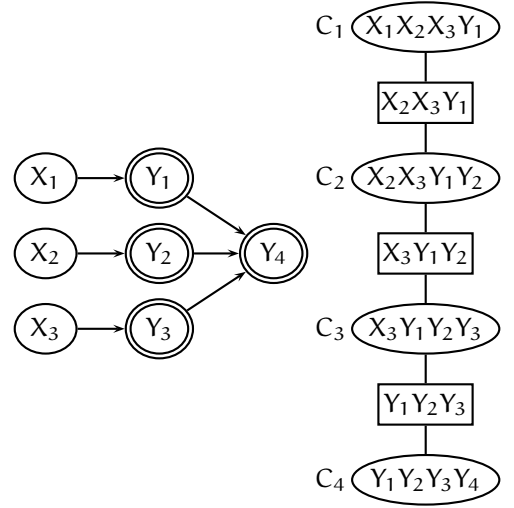

Figure 1: A CLG BN and junction tree

**Example 3.5**
*Figure 1 shows a CLG BN over variables* $Y_i \in \Gamma$ *for* $i =$

$1, \ldots, 4$ and $X_j \in \Delta$ for $j = 1, 2, 3$ and its strong junction tree $\mathcal{T}$. After initialization the clique potentials are:

$$\begin{aligned}
\pi_{C_1} &= (\{P(X_1), P(X_2), P(X_3)\}, \{f(Y_1|X_1)\}), \\
\pi_{C_2} &= (\emptyset, \{f(Y_2|X_2)\}), \\
\pi_{C_3} &= (\emptyset, \{f(Y_3|X_3)\}), \\
\pi_{C_4} &= (\emptyset, \{f(Y_4|Y_1, Y_2, Y_3)\}).
\end{aligned}$$

*The domain of each factor in any clique potential consists of a single variable. This is contrary to both the Lauritzen & Jensen [6] and Cowell [2] architectures where each clique has a probability potential over all discrete variables in the clique. This representation is storage demanding when $Y_4$ has additional parents each having a single discrete variable as parent and when the discrete variables have many states.*

The above example illustrates the structure of a set of CLG BNs used in production by a commercial customer. In this application a large part of the discrete variables are observed making the present inference scheme very efficient on this type of network.

### 3.3 PROPAGATION

Propagation of information in $\mathcal{T}$ proceeds by message passing via the separators $\mathcal{S}$. The separator $S = A \cap B$ between two adjacent cliques $A$ and $B$ stores the messages passed between $A$ and $B$, see Figure 2.

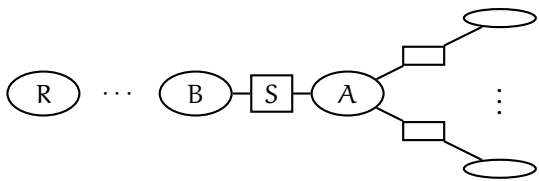

Figure 2: A junction tree with root clique R

Messages are passed from leaf cliques toward R by recursively letting each clique $A$ pass a message to its parent $B$ whenever $A$ has received a message from each $C \in adj(A) \setminus \{B\}$ (COLLECT). Messages are, subsequently, passed in the opposite direction (DISTRIBUTE). DISTRIBUTE is performed from the root to boundary cliques.

### 3.4 MESSAGES

The message $\pi_{A \to B}$ is passed from $A \in \mathcal{C}$ to $B \in adj(A)$ by absorption. Absorption from $A$ to $B$ involves eliminating the variables $A \setminus B$ from the combination of the potential associated with $A$ and the messages passed to $A$ from each neighbor except $B$. The message $\pi_{A \to B}$ is computed as:

$$\pi_{A \to B} = \left(\pi_A \otimes \left(\otimes_{C \in adj(A) \setminus \{B\}} \pi_{C \to A}\right)\right)^{\downarrow B},$$

where $\pi_{C \to A}$ is the message passed from C to A and $\downarrow$ is the projection operation based on EXCHANGE operations and barren variable removals.

### 3.5 THE PUSH OPERATION

A strong junction tree representation $\mathcal{T}$ of a CLG BN is *not always* wide enough to support the insertion of evidence on any continuous variable or the calculation of any posterior marginal density function. If the junction tree is not wide-enough to support a calculation, then the PUSH operation is used [6].

The marginal density of a variable $Y \in \Gamma$ is, in general, a mixture of Gaussian distributions. In order to compute the marginal mixture of Y, it may be necessary to (temporarily) rearrange the content of cliques and separators of $\mathcal{T}$. The PUSH operation is applied in order to rearrange $\mathcal{T}$ such that Y becomes part of a boundary clique. This is achieved by extending cliques and separators to include Y and collecting Y towards R.

Assume $A$ is the clique closest to R such that $Y \in A$, $A \notin bd(\mathcal{C})$, $B = \pi_{\mathcal{C}}(A)$, and $S = \pi_{\mathcal{S}}(A)$, see Figure 2. The PUSH operation extends S and B to include Y. In the process any continuous variable $Z \in T(f)$ such that $Z \notin S$ is eliminated from the density f of Y where $T(f)$ is the tail of f, i.e., the set of conditioning variables. The process of eliminating tail variables not in S is repeated recursively until $T(f) \subseteq S$. The resulting density f is associated with $\pi_B$ and $\pi_{A \to B}$.

The PUSH operation is applied recursively on the parent clique until Y becomes part of a boundary clique.

### 3.6 INSERTION OF CONTINUOUS EVIDENCE

Let $Y \in \Gamma$ be instantiated by evidence $\epsilon_Y = \{Y = y\}$, let $f(Y|\pi(Y))$ be the density function for Y given $\pi(Y)$ and let C be the clique to which $f(Y|\pi(Y))$ is associated. Assume Y has only discrete parents, if any, i.e., $I = \pi(Y) \subseteq \Delta$. Insertion of evidence $\epsilon_Y$ produces a factor $p(y|I)$ such that:

$$p(y|I = i) = \frac{\exp\left(-(y - \alpha_Y(i))^2/(2\sigma^2(i))\right)}{\sqrt{2\pi\sigma^2(i)}},$$

where we assume $\sigma_Y^2(i) > 0$ for all $i$ [6, 2]. The clique potential $\pi_C = (\mathcal{P}, \mathcal{F})$ is updated such that $\pi_C^* = (\mathcal{P} \cup \{p\}, \mathcal{F} \setminus \{f\})$. If $\sigma^2(i) = 0$, insertion of evidence may be undefined, see [2] who cites [6].

If $\pi(Y) \not\subseteq \Delta$, then a sequence of PUSH operations are performed in order to compute the marginal density function for Y. The density f of Y is pushed to the boundary clique. Subsequently, evidence $\epsilon_Y$ is inserted as described above. This implies that the insertion of evidence on a continuous variable may change the content of clique and separator potentials. This occurs when it is necessary to apply the PUSH operation in order to insert $\epsilon_Y$. Finally, Y is instantiated in all density functions where Y is a tail variable.

Notice that $\mathrm{bd}(\mathcal{C})$ may change when $\epsilon_Y$ is inserted.

**Example 3.6**
*Consider the simple CLG BN and its corresponding junction tree $\mathcal{T}$ shown in Figure 3. After initialization the clique potentials are:*

$$\pi_{C_1} = (\{P(X)\}, \{f(Y_1|X)\}),$$
$$\pi_{C_2} = (\emptyset, \{f(Y_2|Y_1)\}).$$

*Assume evidence $\epsilon = \{Y_2 = y_2\}$. Since the tail of $f(Y_2|Y_1)$ is continuous and a subset of the parent separator, it is necessary to apply the PUSH operation on $Y_2$ in order to insert $\epsilon$ into $\mathcal{T}$.*

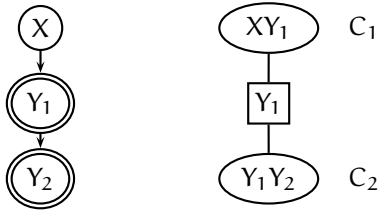

Figure 3: Insertion of evidence on $Y_2$ requires a PUSH operation

*First the density $f(Y_2|Y_1)$ is pushed to the parent clique, next an EXCHANGE operation is performed on the arc $(Y_1, Y_2)$. Next, densities including $Y_2$ in the domain are instantiated to reflect the evidence. Once the PUSH operation completes the revised clique potentials are:*

$$\pi^*_{C_1} = (\{P(X), p(y_2|X)\}, \{f(Y_1|X, y_2)\}),$$
$$\pi^*_{C_2} = (\emptyset, \emptyset).$$

*This completes the insertion of evidence $\epsilon$.*

### 3.7 PROPAGATION OF CONTINUOUS EVIDENCE

Section 3.3 describes the propagation scheme used when $\epsilon_\Gamma = \emptyset$. When $\epsilon_\Gamma \neq \emptyset$, the recursive message passing scheme is terminated at each boundary clique. Once each boundary clique $A \in \mathrm{bd}(\mathcal{C})$ has received messages from each $C \in \mathrm{adj}(A) \setminus \{\pi_\mathcal{C}(A)\}$, continuous evidence is inserted using the PUSH operation.

Let $\mathcal{T} = (\mathcal{C}, \mathcal{S})$ be a strong junction tree representation and let $\epsilon = \epsilon_\Delta \cup \epsilon_\Gamma$ be the evidence to propagate. The evidence $\epsilon$ is propagated in $\mathcal{T}$ by performing the following sequence of steps:

1. Initialization including insertion of evidence $\epsilon_\Delta$.
2. At each $A \in \mathrm{bd}(\mathcal{C})$ COLLECT from every $B \in \mathrm{adj}(A) \setminus \{\pi_\mathcal{C}(A)\}$.
3. Insert evidence $\epsilon_\Gamma$ using the PUSH operation.
4. Perform in sequence a COLLECT and a DISTRIBUTE operation on R.

During the COLLECT operation performed in step 4 messages are passed from the boundary cliques to R. Thus, the effect of steps 2 and 4 is that two messages have been passed between each pair of adjacent cliques on any path between the root R and a boundary clique. No messages are passed from boundary cliques to leave cliques.

The architectures described in [2], [6], and [9] each does a full propagation over all the nodes of the computation tree prior to inserting $\epsilon$ whereas we do only a partial COLLECT prior to inserting $\epsilon_\Delta$.

### 3.8 POSTERIOR MARGINALS

The posterior marginal $P(X|\epsilon)$ for $X \in \Delta$ may be computed from any clique or separator containing X. Since $\epsilon_\Gamma$ is incorporated using PUSH operations, no $Y \in \Gamma$ needs to be eliminated in the process of computing $P(X|\epsilon)$. If $X \in C$, then $P(X|\epsilon)$ is computed as:

$$P(X|\epsilon) \propto \sum_{C \setminus \{X\}} c(\pi_C) = \sum_{C \setminus \{X\}} \prod_{p \in \mathcal{P}_C} p \cdot \prod_{f \in \mathcal{F}_C} f$$
$$= \sum_{C \setminus \{X\}} \prod_{p \in \mathcal{P}_C} p,$$

where $\pi_C = (\mathcal{P}_C, \mathcal{F}_C)$ is the clique potential for C. On the other hand, if S is a separator containing X with adjacent cliques A and B, then $P(X|\epsilon)$ is computed as:

$$P(X|\epsilon) \propto \sum_{S \setminus \{X\}} c(\pi_{A \to B} \otimes \pi_{B \to A})$$
$$= \sum_{S \setminus \{X\}} \prod_{p \in \mathcal{P}_{A \to B} \cup \mathcal{P}_{B \to A}} p \cdot \prod_{f \in \mathcal{F}_{A \to B} \cup \mathcal{F}_{B \to A}} f$$
$$= \sum_{S \setminus \{X\}} \prod_{p \in \mathcal{P}_{A \to B} \cup \mathcal{P}_{B \to A}} p,$$

where potential $\pi_{A \to B} = (\mathcal{P}_{A \to B}, \mathcal{F}_{A \to B})$ and potential $\pi_{B \to A} = (\mathcal{P}_{B \to A}, \mathcal{F}_{B \to A})$ are the potentials passed from A to B and from B to A, respectively.

The posterior mixture for $Y \in \Gamma$ is computed using PUSH operations followed by a projection of the relevant boundary clique to Y and a contraction.

**Example 3.7**
*The prior mixture densities of $Y_1$ and $Y_2$ of the CLG BN shown in Figure 4 are:*

$$f(Y_1) = \sum_{x_1 \in X_1} P(x_1) \cdot f(Y_1|x_1),$$
$$f(Y_2) = \sum_{x_1 \in X_1, x_2 \in X_2} P(x_1) P(x_2|x_1) \cdot f(Y_2|x_1, x_2).$$

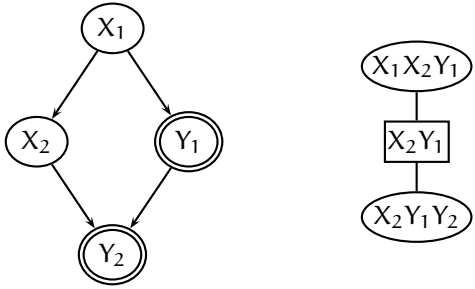

Figure 4: Prior density for $Y_1$ has $\|X_1\|$ components

*The density for $Y_1$ has only $\|X_1\|$ components. This is a reduction compared to the Lauritzen & Jensen and Cowell architectures where the marginal density will have $\|X_1\| \cdot \|X_2\|$ components. The reduction is due to the decomposition of clique and separator potentials.*

**Example 3.8**
*Consider again Figure 1 of Example 3.5. The number of components in the mixture marginal for $Y_4$ is $\|X_1\| \cdot \|X_2\| \cdot \|X_3\|$ whereas the number of components in the mixture marginal for $Y_i$ is equal to $\|X_i\|$. This is a reduction compared to the Lauritzen & Jensen and Cowell architectures where the number of components is $\|X_1\| \cdot \|X_2\| \cdot \|X_3\|$. Hence, in the case of a larger number of variables (and same structure), the storage and time reduction can be significant.*

## 4 COMPARISON

### 4.1 COWELL

Cowell [2] presents an algorithm for belief update where message passing proceeds on an elimination tree rather than a (strong) junction tree. This produces a local propagation scheme in which all calculations involving continuous variables are performed by manipulating univariate regressions (avoiding matrix operations) such that continuous variables are eliminated using EXCHANGE operations.

The three main differences between the present propagation scheme and Cowell [2] are: use of a strong junction tree as opposed to an elimination tree, use of EXCHANGE to eliminate both continuous and discrete variables and a single round of message passing.

### 4.2 LAURITZEN AND JENSEN

The architecture of Lauritzen & Jensen [6] performs belief update by message passing in a strong junction tree architecture. A *CG potential* [8] is associated with each clique and separator. A CG potential consists of a probability potential over discrete variables and a probability density function over continuous variables conditional on the discrete variables. Each clique and separator has a CG potential over its variables. This implies that complex matrix operations are required during belief update.

Initialization plays an important role in the Lauritzen & Jensen [6] architecture. It produces a Lauritzen & Spiegelhalter-like junction tree representation [7] where clique potentials are conditioned on the continuous variables of the parent separator. This ensures that a variable $Y \in \Gamma$ is only propagated when inserting evidence on Y or when computing the mixture marginal for Y. Furthermore, a complex recursive combination operator may be required during initialization in order to combine CG potentials. The need for conditioning, recursive combination, and complex matrix operations is eliminated in both the Cowell [2] and the present architectures.

**Example 4.1**
*Figure 5 shows a CLG BN and its junction tree $\mathcal{T}$. The initial clique potentials are:*

$$\pi_{C_1} = (\{P(X_1)\}, \{f(Y_1|X_1), f(Y_3|X_1, Y_1, Y_2)\}),$$
$$\pi_{C_2} = (\emptyset, \{f(Y_2|Y_1, Y_4), f(Y_4)\}).$$

*In the Lauritzen & Jensen [6] architecture initialization of $\mathcal{T}$ requires a recursive combination operation.*

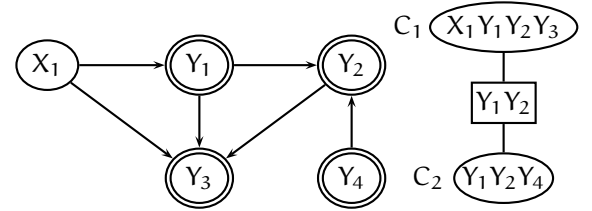

Figure 5: Initialization requires recursive combination in the Lauritzen & Jensen architecture

*In the proposed architecture initialization amounts to associating probability distributions and densities with cliques of $\mathcal{T}$. The prior distribution of each variable is readily computed using the EXCHANGE operation.*

The Lauritzen & Jensen [6] architecture calculates weak marginals during DISTRIBUTE. This is not the case for the Cowell [2] nor the present architecture.

### 4.3 MADSEN

The present architecture is quite different from the architecture proposed by Madsen [9]. The latter architecture is an extension of Madsen & Jensen [11] to the case of CLG BNs based on the propagation scheme of Lauritzen & Jensen [6]. This implies a number of differences when compared to the present scheme. First, the architecture is based solely on the operations of Lauritzen & Jensen [6] whereas the present scheme is based on operations of both Lauritzen & Jensen [6] and Cowell [2]. Second, a Lauritzen & Spiegelhalter-like junction tree representation is achieved as the result of initialization, i.e., during the initial COLLECT operation, the sender clique is conditioned on the continuous variables of the parent separator. Finally, in the present scheme variable eliminations are performed using EXCHANGE operations and barren variable removals.

## 5 PERFORMANCE ANALYSIS

A preliminary performance analysis on a set of randomly generated CLG BNs has been made. Networks with 25, 50, 75, 100, 125, and 150 variables with different fractions of continuous variables (0, 0.25, 0.5, 0.75, 1) were randomly generated (ten networks of each size). For each network, evidence sets of 0 to 20 instantiated variables were generated and 40 sets of evidence were generated for each size.

We compared the performance of the present architecture with the performance of the commercial implementation of the Lauritzen & Jensen [6] architecture in the HUGIN Decision Engine. Table 1 shows statis-tics on one of the networks used in the tests (net50-4) where $s(C) = \prod_{X \in \Delta \cap C} \|X\|$ and $s(\mathcal{C}) = \sum_{C \in \mathcal{C}} s(C)$. Figure 6 shows the average time cost of belief update in net50-4 whereas Figure 7 shows the average size of the largest discrete configuration. A discrete configuration is either the domain of a factor or the discrete conditioning set of a density. This is an example where the proposed architecture is most efficient.

Table 1: Statistics On CLG BN net50-4

| Network | $|\mathcal{X}|$ | $|\mathcal{C}|$ | $\max_{C \in \mathcal{C}} s(C)$ | $s(\mathcal{C})$ |
|---|---|---|---|---|
| net50-4-0 | 50 | 42 | 3,888 | 18,084 |
| net50-4-0.25 | 50 | 39 | 186,624 | 231,309 |
| net50-4-0.5 | 50 | 38 | 165,888 | 218,656 |
| net50-4-0.75 | 50 | 39 | 1,728 | 2,444 |
| net50-4-1 | 50 | 40 | 1 | 40 |

The present architecture maintains a factorization of clique and separator potentials into sets of factors and densities. This decomposition implies that the largest discrete domain size considered during belief update is often significantly smaller than the discrete domain size of the largest clique in the strong junction tree. This insight is supported by the experimental analysis, which indicates that the Lauritzen & Jensen [6] implementation runs out of memory on most networks with 75 or more variables for a large fraction of the evidence sets whereas the present architecture runs out of memory on a much smaller fraction of the evidence sets. Figure 7 illustrates how the average largest discrete domain size decreases as $|\epsilon|$ increases. Notice that the average largest size is significantly smaller than the size of the largest clique in the strong junction tree.

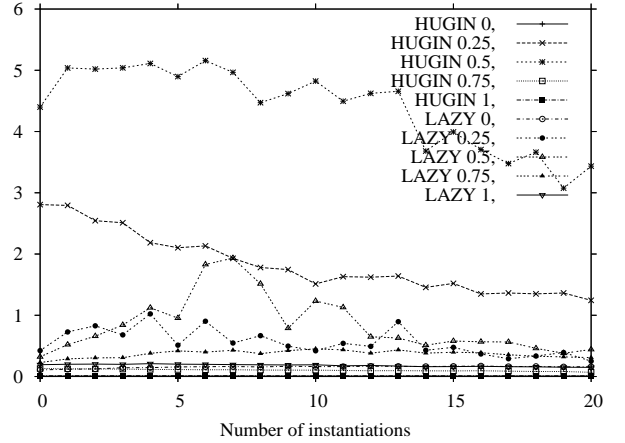

Figure 6: Average time in sec. for belief update

For networks with only discrete or only continuous variables the Lauritzen & Jensen implementation is faster than the implementation of the proposed architecture. However, for some networks with a fraction of 0.25 or 0.5 continuous variables Lauritzen & Jensen is significantly slower than the proposed architecture.

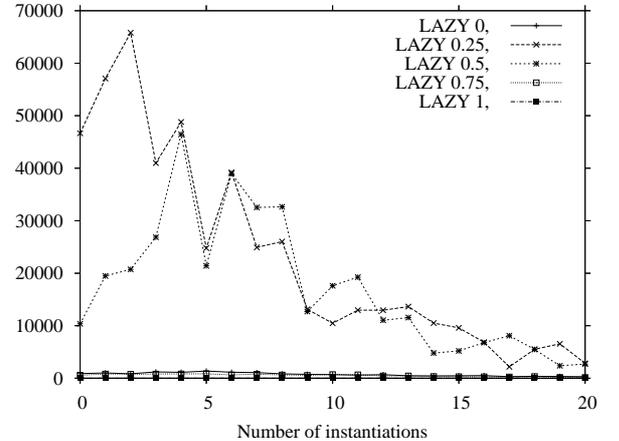

Figure 7: Average size in numbers

The typical decrease in average time cost as $|\epsilon|$ increases for lazy propagation is not as significant on CLG BNs. The reason is that computing marginal densities is a dominant and a non-constant factor in the time cost of belief update. A significant amount of the total time for propagating evidence is spent on computing posterior mixture marginals. In the proposed architecture the number and the computational cost of PUSH operations is reduced by a decomposition of clique and separator potentials. The significance of the decrease depends on the ratio of continuous variables. Figure 8 shows the average time cost of computing marginals in net50-4. Notice that a significant amount of the time cost originates from computing marginals.

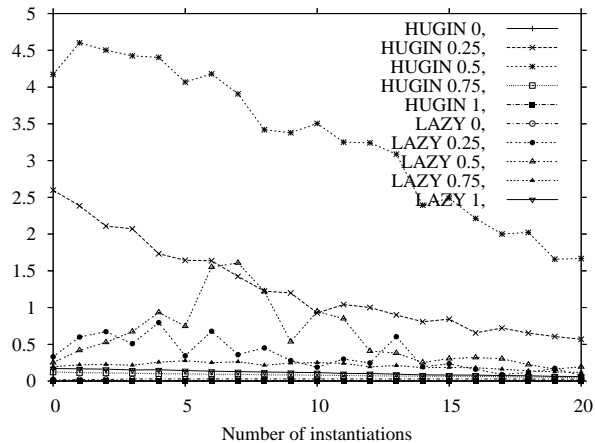

Figure 8: Average time in sec. for marginals

On most of the networks considered in the test — where belief update is feasible — the commercial implementation of Lauritzen & Jensen [6] is most efficient (typically networks with less than 75 variables). The HUGIN Decision Engine has significantly more efficient data structures and operations than the implementation of the proposed architecture though.

The experiments were performed using a Java implementation on a desktop computer with a 2.2 GHz AMD Athlon$^{\text{TM}}$ CPU and 768 MB RAM running Redhat 8.

## 6 CONCLUSION

An architecture for belief update in CLG BNs based on lazy propagation where messages are computed using EXCHANGE operations and barren variable eliminations has been presented. The architecture is based on extended versions of operations introduced by Lauritzen & Jensen [6] and Cowell [2].

Despite a significant difference in the efficiency of table operations the proposed architecture is — in some cases — more efficient than a commercial implementation of the Lauritzen & Jensen [6] architecture. The results of the performance evaluation indicate a significant potential of the proposed architecture.